%% file: PaperForReview.tex
\documentclass[10pt,twocolumn,letterpaper]{article}

\usepackage{wacv}              

\usepackage{graphicx}
\usepackage{amsmath}
\usepackage{amssymb}
\usepackage{booktabs}

\usepackage{amssymb}
\usepackage{pifont}
\newcommand{\cmark}{\ding{51}}%
\newcommand{\xmark}{\ding{55}}%
\usepackage[pagebackref,breaklinks,colorlinks]{hyperref}

\usepackage{graphicx}
\usepackage{amsmath}
\usepackage{amssymb}
\usepackage{booktabs}

\usepackage{times}
\usepackage{epsfig}

\usepackage{color}
\usepackage{tabularx}

\usepackage{times}
\usepackage{epsfig}
\usepackage{graphicx}
\usepackage{amsmath}
\usepackage{amssymb}

\usepackage{booktabs}
\usepackage{colortbl}
\usepackage{url}
\usepackage{wrapfig}
\usepackage{blindtext}

 \usepackage{multirow}
\usepackage[utf8]{inputenc} 
\usepackage[T1]{fontenc}    
\usepackage{amsfonts}       
\usepackage{nicefrac}       
\usepackage{microtype}      
\usepackage{xcolor}  
\usepackage{diagbox}

\usepackage{marvosym} 

\usepackage{fontawesome} 

\usepackage[accsupp]{axessibility} 

\usepackage[pagebackref,breaklinks,colorlinks]{hyperref}

\usepackage[capitalize]{cleveref}
\crefname{section}{Sec.}{Secs.}
\Crefname{section}{Section}{Sections}
\Crefname{table}{Table}{Tables}
\crefname{table}{Tab.}{Tabs.}

\usepackage{algorithm}
\usepackage{algorithmic}
\usepackage{times}

\def\wacvPaperID{1955} 
\def\confName{WACV}
\def\confYear{2024}

\begin{document}

\title{Multi-Modal Gaze Following in Conversational Scenarios}



\author{
Yuqi Hou\textsuperscript{1} \quad Zhongqun Zhang\textsuperscript{1} \quad Nora Horanyi\textsuperscript{1} \quad Jaewon Moon\textsuperscript{2} \quad \\Yihua Cheng\textsuperscript{\faEnvelopeO 1} \quad Hyung Jin Chang\textsuperscript{1}\\
\textsuperscript{1}University of Birmingham, United Kingdom \quad
\textsuperscript{2}KETI, Korea\\
{\tt\small \{yxh029,zxz064\}@student.bham.ac.uk, nxn840@alumni.bham.ac.uk, jwmoon@keti.re.kr}\\
{\tt\small \{y.cheng.2,h.j.chang\}@bham.ac.uk}
}


\maketitle

\begin{abstract}
    Gaze following estimates gaze targets of in-scene person by understanding human behavior and scene information. 
    Existing methods usually analyze scene images for gaze following. 
    However, compared with visual images, audio also provides crucial cues for determining human behavior.
    This suggests that we can further improve gaze following considering audio cues. 
    In this paper, we explore gaze following tasks in conversational scenarios. 
    We propose a novel multi-modal gaze following framework based on our observation ``audiences tend to focus on the speaker''. 
    We first leverage the correlation between audio and lips, and classify speakers and listeners in a scene. We then use the identity information to enhance scene images and propose a gaze candidate estimation network. The network estimates gaze candidates from enhanced scene images and we use MLP to match subjects with candidates as classification tasks.
    Existing gaze following datasets focus on visual images while ignore audios.
    To evaluate our method, we collect a conversational dataset, VideoGazeSpeech (VGS), which is the first gaze following dataset including images and audio. 
    Our method significantly outperforms existing methods in VGS datasets. The visualization result also prove the advantage of audio cues in gaze following tasks. Our work will inspire more researches in multi-modal gaze following estimation.
\end{abstract}

\section{Introduction}
\label{sec:intro}

Human gaze provides important cues for understanding human behavior and is required by various fields such as social communication\cite{masse2017tracking} and human-robot interaction\cite{marin2014detecting}. Gaze following is a crucial topic in gaze estimation. It provides human intention in one scene and demanded by human-robot interaction\cite{mcmillan2019designing}.

Gaze following aims to estimate gaze targets in a scene where the human subjects appear in the same scene. Existing researches\cite{sumer2020attention,fan2019understanding,epstein2020oops,wei2018looking,chong2020detecting,Li_2021_ICCV} usually leverage facial and scene information to estimate the gaze target. Compared with vision information, audio also provides important cues in a scene. 
Previous studies \cite{tavakoli2019dave, van2008audiovisual, doyle2001identification} have demonstrated the important role of sound in audiovisual estimation. 
Numerous psychological studies also \cite{van2008pip, vroomen2000sound, perrott1990auditory, vroomen2000sound, min2014sound} have highlighted the significant impact of sound on visual attention. 
Combining audio and vision modalities can provide richer and complementary information compared to unimodal approaches \cite{gao2020survey}. 
However, to the best of our knowledge, there is no research using audio information in gaze following.
\input{fig/teaser/item.tex}

Conversational scenarios are among the most common scenarios, emerging whenever there are more than two individuals present in a scene.
In this work, we explore gaze following in conversational scenarios. 
We propose a multi-modal gaze following framework (MMGaze) which leverages both visual and audio cues.
The framework is designed based on our observation, \textit{``audiences tend to focus on the speaker''}.
It means the gaze following performance would be naturally improved if we add identity information for gaze target inference.
Therefore, our framework first performs active speaker detection based on the correlation between lip motion and audio. 
We respectively perform face tracking in videos and the mel-frequency cepstrum coefficients (MFCC) feature\cite{davis1980comparison} extraction from the audio.
We compare lip motion features of each individual with MFCC features to distinguish speaker and listener~\cite{chung2016out}.
To add identity information into scene images, we respectively generate binary identity maps for speakers and listeners. We stack the two maps with scene images for scene image enhancement. 

We further build a gaze candidate estimation network which predicts all gaze candidates from enhanced scene images. The network is inspired by object detection tasks which detect objects from scene images.
We use one MLP to match subjects with gaze candidates.
The MLP performs binary classification tasks and we select the candidate with the largest probability as the final gaze target for one subject.
Existing gaze following datasets usually focus on visual images.
We collect a new gaze following dataset, VideoGazeSpeech (VGS), to evaluate MMGaze.
We manually annotate the dataset and require three different reviewers to check the annotation, which ensures the correctness of our dataset.
VGS contains $29$ videos with audio tracks consisting of $35,231$ frames.
To facilitate future research, we also provide annotations in different formats, including VOC format, COCO format, and VideoAttentionTarget\cite{chong2020detecting} format. 

The main contributions of our work are threefold.

\begin{itemize}
    \setlength{\itemsep}{2pt}%
    \setlength{\parskip}{2pt}%
    \vspace{-2mm}

    \item[$\bullet$] We propose the MMGaze for multi-modal gaze following. The framework predicts identity information based on the correlation of lip motion and audio. We employ the identity information to enhance scene images and propose a gaze candidate network which estimates all gaze candidates from enhanced images.

    \item[$\bullet$] To evaluate our method, we introduce a new gaze following dataset, which is also the first gaze following dataset containing audio track. Our dataset would encourage future research in multi-modal gaze following.
    
    \item[$\bullet$]  We evaluate our method on the VGS dataset. Our method has the best performance and experiments demonstrate the advantage of audio in gaze following.
                           
\end{itemize}

\input{tbl/quantitative/item2.tex}
\section{Related Works}

Despite the significance of this topic, research in this area is surprisingly limited. To address this gap, the project divides the work into three parts. The first part provides an overview of the existing research on gaze target detection. The second part reviews the current state of research on multimodal approaches to gaze detection. Finally, the third part presents a list of available gaze datasets that can be used to train and evaluate gaze detection models.

\subsection{Gaze Following Methods}

Gaze tracking is a crucial area of research in computer vision, with numerous applications in fields such as human-computer interaction and medical diagnosis~\cite{Cheng_2021_arxiv,cheng_2022_aaai,Bao_2020_ICPR,Cheng_2023_ICCV}. However, existing gaze following methods often rely on traditional gaze tracking devices\cite{Cheng_2020_AAAI,Cheng_2018_ECCV,horanyi2023g,cheng2022icpr}, which can interfere with the user's natural gaze behavior and limit the accuracy of results. 
To overcome this limitation, recent studies explore deep learning for gaze following in images or videos.

Current gaze following research is limited to image format as input and does not consider audio information. Most studies obtain the prediction results of gaze following target by combining raw frame with head position input \cite{recasens2015they,chong2020detecting,fang2021dual,hu2022we,Horanyi_2023_CVPR}. Recently, Tu et al. \cite{tu2022end} redefined the gaze following task to predict the paired head position and gaze target by inputting raw frames. While these approaches are limited to the analysis and learning of image information, our daily activities, based on many psychological experiments, suggest that our gaze following relies not only on visual senses but also on auditory information \cite{van2008pip,vroomen2000sound,perrott1990auditory,min2014sound}. Tavakoli et al. \cite{tavakoli2019dave} proposed that audio signal contributes significantly to dynamic saliency prediction.

Furthermore, most existing deep learning prediction models for gaze following require the input of head position along with raw frames \cite{recasens2015they,chong2020detecting,fang2021dual,hu2022we}, which is inconvenient for the flexible application of the model. Additionally, adding head position as input to the expected picture would be redundant and impractical. 
In contrast, our proposed network architecture only requires raw video input. It can automatically predict the head position of both speaker and listener using contrastive learning and directly output the gaze following target. This network does not require additional input of head position in the input part and is, therefore, a more flexible and convenient end-to-end process.

\subsection{Multimodal}
As for multimodal research, there is surprisingly a few multimodal vision research for the gaze following domain \cite{hu2022we,Nonaka_2022_CVPR}. 
So this work aptly fills this gap by innovatively merging audio and video information.

Nonaka et al.\cite{Nonaka_2022_CVPR} formulated gaze estimation as Bayesian  prediction, rather than an artificial way, where they estimate the likelihoods of head and body orientations given an input image, and then multiply a learned conditional temporal prior of gaze direction by cascading two neural networks. Hu et al. \cite{hu2022we} propose a novel extension method that adds 3D space by the use of depth information, which is not strictly multimodal fusion application.


Besides these gaze works, the multimodal task in computer vision is currently performed by two main factions, Fuse and Align. The Fuse faction fuses in a single tower structure, and this faction mainly applies the Transformer. The Transformer's attention has the ability to aggregate features in different feature spaces and at a global scale. The Transformer is suitable for alignment and fusion of multimodal feature representations. Vision Transformer \cite{dosovitskiy2020image} was proposed to break the model barrier between CV and NLP. The Align faction of fusion is a two-tower structure, represented by CLIP \cite{radford2021learning} and ALIGN \cite{jia2021scaling}, focusing on multimodal alignment for downstream tasks such as graphical matching and retrieval. The VGS structure proposed in this paper is a multimodal fusion approach based on the latter.

Furthermore, multimodal research in computer vision has intensified in recent years in terms of the classes of elements combined in modality\cite{vasudevan2018object}, combines language and gaze and proposes the object referring dataset and framework that the observer is watching while describing and watching the video. Boccignone et al.\cite{boccignone2020gaze} mentioned the spatial-temporal multimodal input that fuses audio and video and applies the Foraging framework. D'Amelio and Boccignone \cite{d2021gazing} improved the way of weighing the patch in the Foraging framework. Nevertheless, they detected the eye-tracking data of the viewer watching the video, not the gaze-following target of the people in the video.
All these methods are good at fusing multiple modalities, but they do not explore the role of sound as an aid to gaze estimation. In light of this, this thesis will close a research gap in audio-video fusion in the realm of gaze following.



\subsection{Gaze Datasets}
A summary of comparable gaze datasets is shown in Table \ref{table1}.
Publicly available datasets for gaze estimation typically focus on in-the-wild scenes or video programs and currently only have visual unimodality. These datasets can be classified based on various factors, such as dimension (2D or 3D), format (video or image), frame type (in, out, or cross), annotation method (gaze direction or gaze target), and modality (vision or audio-vision).

For 2D datasets, GazeFollow \cite{recasens2015they} marks the center of a person's eyes and where they are looking with only in-frame annotations, disregarding out-of-frame cases. The VideoGaze dataset \cite{recasens2016following} is a large dataset for gaze tracking across multiple views, but it requires pairing frames individually. The VideoAttentionTarget \cite{chong2020detecting} includes $109,574$ in-frame and out-of-frame fixation comments and $54,967$ comments, but it only labels the classification for out-of-frame images, ignoring the target ground truth. VideoCoAtt \cite{fan2018inferring} is a dataset of $380$ complex video sequences from public TV shows, specifically designed for shared attention research.For 3D datasets, Gaze360 \cite{kellnhofer2019gaze360} is a 3D gaze tracking dataset that includes subjects in indoor and outdoor environments, labeled with 3D gaze at various head poses and distances. The RGB-D attention dataset \cite{hu2022we} contains everyday human activities with 3D gaze target annotations. The GazeFollow360 dataset \cite{Li_2021_ICCV} collected videos into 360-degree images in the equirectangular format.

While these datasets provide a good starting point for gaze estimation research, more diverse and comprehensive datasets are still needed to better capture the complexities of real-world gaze estimation, including out-of-frame gaze estimation, diverse scenarios, and multimodal input. Our proposed VGS dataset addresses these issues by focusing on conversational scenarios, overcoming the out-of-frame issue, and fusing audio cues to improve the diversity and robustness of gaze target detection.

\begin{algorithm}[t]
\caption{Multimodal Gaze Target Detection}
\label{alg:multimodalGaze}
\begin{algorithmic}[1]
\REQUIRE Video stream $V$, Audio signal $A$
\ENSURE Gaze targets $G$

\STATE \textbf{Initialization:}
\STATE \quad Load models: 
\STATE \quad $syncNet$\cite{chung2016out}, $s3fd$\cite{zhang2017s3fd}, $ resneXt$, $rpn$
\STATE \quad Initialize operations: $roiAlign$, $fcn$, $mlp$


\STATE \textbf{Active Speaker Detection:}
\FOR{$frame$ in $V$}
\STATE \quad $face \leftarrow s3fd.detect(frame)$
\STATE \quad Store detected face for timeline creation
\ENDFOR

\STATE Extract audio features $mfcc$ and compute correspondence score
\STATE Identify $speaker$ and $listener$ based on correspondence

\STATE \textbf{Gaze Candidate Estimation:}
\STATE Construct identity maps using $bbox_s^i$ and $bbox_l^i$
\STATE $featureMaps \leftarrow resNext(F)$

\FOR{$point$ in $featureMaps$}
\STATE \quad Define and classify $roi$ using anchors and $rpn$
\STATE \quad Refine $candidateROIs$ with $roiAlign$
\STATE \quad Generate $mask$ using $fcn$
\STATE \quad Predict $gazeTarget$ using $mlp$
\STATE \quad Store $gazeTarget$
\ENDFOR

\RETURN Gaze targets $G$
\end{algorithmic}
\end{algorithm}


\input{fig/framework/item.tex}

\section{Multi-Modal Gaze Following Framework}
\subsection{Overview}
In this paper, we explore gaze following in conversational scenarios.
We propose a multi-modal gaze following framework (MMGaze) to integrate vision and audio information in the scenario.
MMGaze is built based on our key observation ``the audience more likely looks at speakers''.
As shown in Fig.\ref{framework}, MMGaze first performs active speaker detection in scene images.
We leverage the correlation between audio and lips~\cite{chung2016out}, and classifies speakers and listeners.
We then utilize the identity information to enhance scene images.
We propose a gaze candidate estimation network which predicts all gaze candidates from enhanced scene images. A detailed step-by-step process is presented in Algorithm~\ref{alg:multimodalGaze}.

\subsection{Active Speaker Detection}
In daily interactions, speakers naturally attract a greater amount of attention from their audiences. This phenomenon can also extend to conversational scenarios, where audiences tend to focus their gaze on speakers. This observation motivates us to integrate the identity information for gaze following task.
In this work, we distinguish the identity information via multi-modal cues. 
Our work leverages the correlation between audio and lip motion~\cite{chung2016out} to detect active speakers.
We respectively obtain audio features and visual features corresponding to the lip motion of each individual.
We compute the similarity between the two features and distinguish speakers based on the similarity.
A threshold is set to avoid out-of-frame voices.

In detail, we first split the input video frame by frame for face detection.
We use S3FD \cite{zhang2017s3fd} to obtain gray-scale facial images and crop mouth region based on facial landmarks. We stack every five consecutive frames for speaker detection. The sample frequency of videos are $25$ fps where five frames equals to a $200$ ms videos. 
On the other hand, we use 13-dimensional MFCC feature to represent audio cues. The audio is sampled at $100$ HZ. We obtain audios of $20$ frames, which is equal to $200$ ms audio, to align with the video.
To obtain the similarity between lip motion and audios, we use SyncNet~\cite{chung2016out} to extract lip motion feature and audio features.
The SyncNet is trained with contrastive loss which requires lip motion feature should be similar with the corresponding audio feature.
It uses Euclidean distances to measure the similarity of two features.
In our work ,we also uses the distance for speaker detection, where the highest correspondence indicates the speaker.
To avoid out-of-frame voice, we empirically set a threshold for the speaker.

Overall, we leverage the correlation between lip motion and audios for speaker detection. We crop lip region of each individual from scene images and distinguish their identity.
As the result, we have the facial bounding box $\{bbox_s^i\}$ of speakers and $\{bbox_l^i\}$ of listener.

\subsection{Gaze Candidate Estimation}
In this section, we enhance scene images with identity and estimate gaze candidate from the enhanced images.
We then match subjects with candidates via a MLP.

Our work has facial bounding boxes $\{bbox_s^i\}$ and $\{bbox_l^i\}$ based on active speaker detection.
We convert these bounding boxes into identity maps to enhance scene images.
In detail, we construct two identity maps representing speakers and listeners.
The identity map is a binary image and has the same size as scene images.
We mark the facial region of speakers and listeners based on $\{bbox_s^i\}$ and $\{bbox_l^i\}$.
We stack the two identity maps with scene images in the channel dimension.
The five-channel image is used for gaze candidate estimation next.

We propose a gaze candidate estimation network for target detection through supervised learning, which is inspired by object detection task~\cite{he2017mask}. 
The network regresses bounding box of gaze candidate from enhanced scene images.
The process can be broken down into the following steps:
To detect gaze targets, our model inputs enhanced images into ResNeXt101 model\cite{xie2017aggregated} to obtain corresponding feature maps. 
We then set predetermined regions of interest (ROIs) for each point in the feature map using anchors, which gives us multiple candidate ROIs. These candidate ROIs are sent to the Region Proposal Network (RPN) for binary classification and bounding-box regression. We filter out some candidate ROIs and refine the remaining ones using the ROIAlign\cite{he2017mask}, which maps the original image to the corresponding pixels in the feature map and produces a fixed-size feature map.
We introduce candidate frame regression to these ROI regions and use a fully convolutional network (FCN) to generate a mask, which completes the target detection task and outputs all the gaze candidates of all the subjects inside the frame. We final train a Multi-Layer Perceptron (MLP) to map each subject and their gaze targets with the highest probability.


\begin{figure*}[h]
\centering
\includegraphics[width=1.0\linewidth]{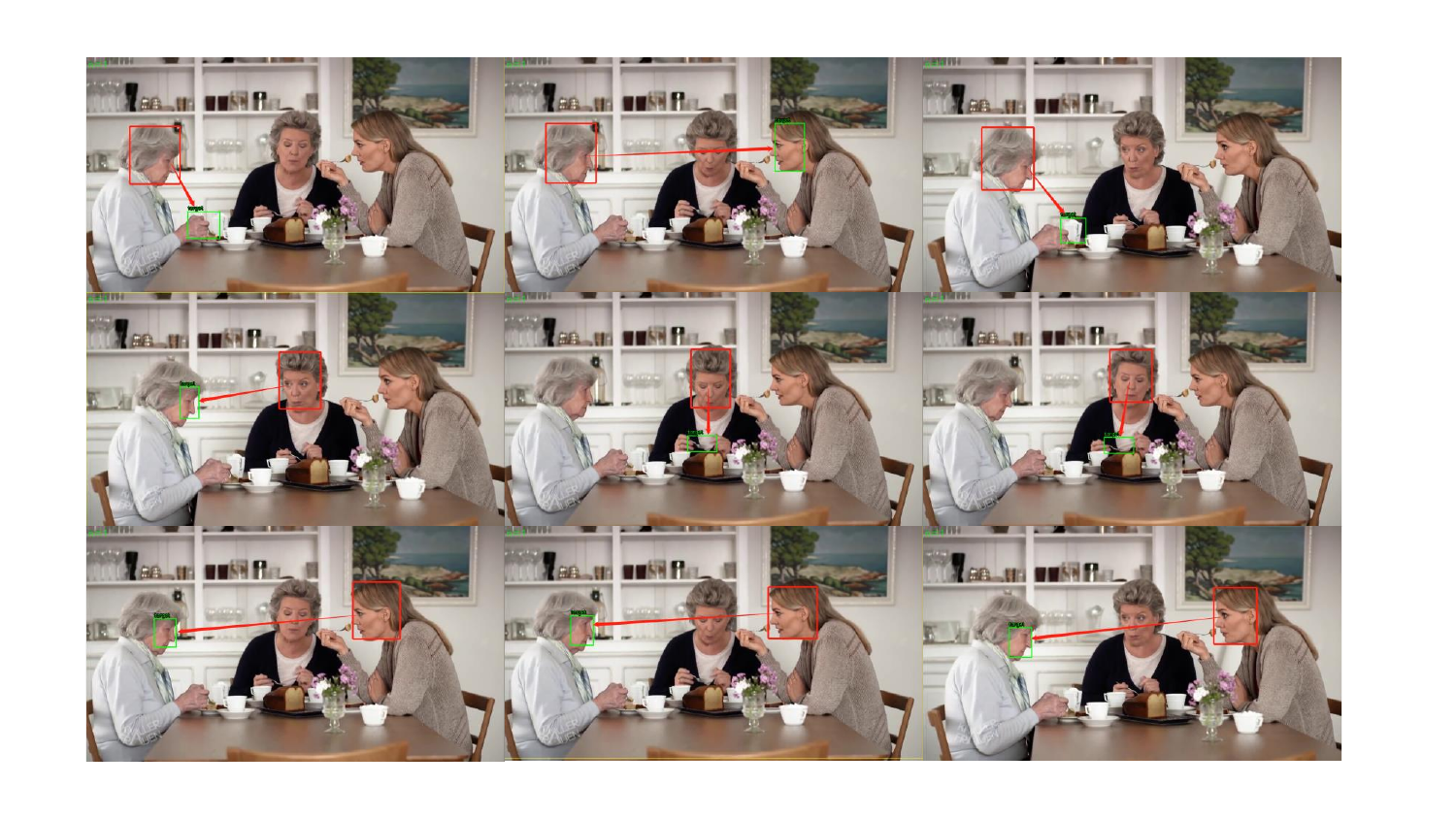}
\vspace{-0.6cm}
\caption{\textbf{Example diagram of the VideoGazeSpeech (VGS) database.}There are three people in the sample video. Each line in the above figure is labelled with the gaze following each person in the video, with the green box indicating the gaze following the target and the red box indicating the corresponding head of the person producing the gaze work}
\vspace{-0.3cm}
\label{fig:fig3}
\end{figure*} 

\subsection{Objective Function}
We use binary cross-entropy loss $\mathcal{L}_{\text{cls}}$ for matching subjects with gaze candidates, and smooth L1 loss for the bounding box prediction of gaze candidates. The smooth L1 loss is shown as follows:
\begin{equation}
\label{Lbox}
     \mathcal{L}_{\text{bbox}}(t, \hat{t}) = \sum_{i \in \{x,y,w,h\}} \mathcal{L}_{smooth}(t_i - \hat{t_i})
,
\end{equation}
Denote predicted box parameterized by $t$ and ground-truth $\hat{t}$, the discrepancy between these two representations is quantified using $\mathcal{L}_{smooth}$. 

\begin{equation}
\label{smooth}
\mathcal{L}_{smooth}(x) =
\begin{cases} 
    0.5x^2 & \text{if } |x| < 1 \\
    |x| - 0.5 & \text{otherwise}
    \end{cases}
,
\end{equation}
The \(\text{smooth}_{L1}\) function is a robust loss function, which serves to mitigate the influence of outliers. For values of \( x \) less than 1, it defaults to an \( L_2 \) loss, ensuring that it is smooth near zero. However, for larger values of \( x \), it behaves linearly, akin to the \( L_1 \) loss, thus ensuring robustness against larger deviations. This amalgamation of \( L_1 \) and \( L_2 \) characteristics makes it particularly suitable for regression tasks in the presence of potential outliers.

\section{VideoGazeSpeech Dataset}
Gaze following attracts much attention recently \cite{chong2020detecting,recasens2015they} while existing databases commonly lack audio information. In this work, we collect the first gaze following dataset containing audios, the VideoGazeSpeech Dataset. The dataset is used to evaluate our method and also encourage future research in multi-modal gaze following. 
Samples from our dataset are presented in Fig.\ref{fig:fig3}. 
Our dataset comprises a total of $35,231$ frames of $29$ videos. Each video in the dataset has an average duration of approximately $20$ seconds and is recorded at a frame rate of $25$ frames per second (fps). The resolution of each video is $1280 \times 720$ pixels, and the entire dataset occupies a storage space of $7.2$ GB.

\begin{figure*}[t]
\centering
\includegraphics[width=1.0\linewidth]{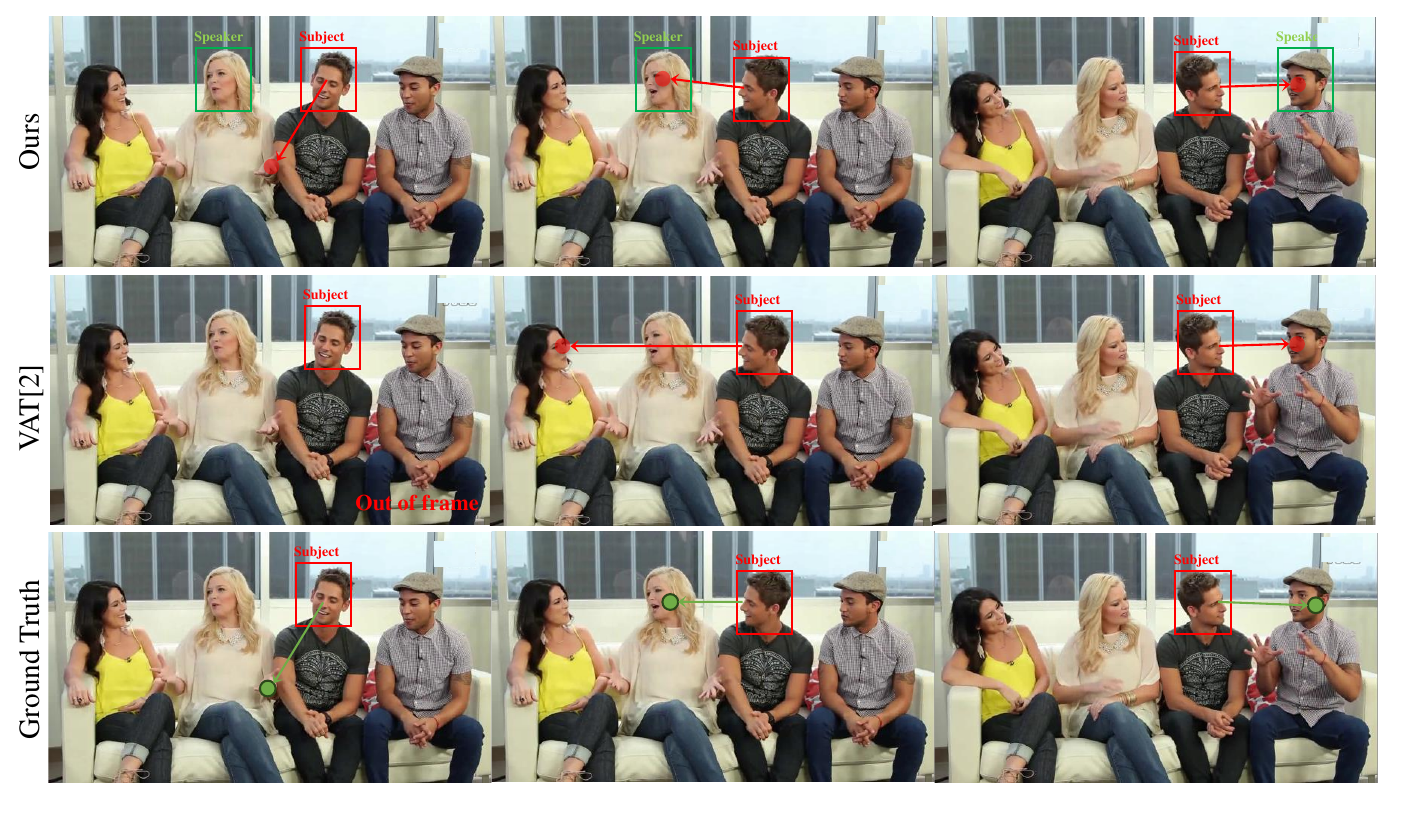}
\vspace{-0.6cm}
\caption{\textbf{Comparison from Gaze Candidate Estimation model and VAT model. }The first line is the output of gaze candidate estimation model, the second line is the output of VAT model, and the third line is the ground truth. Our model outperforms the VAT method in accurately detecting the gaze target. In the first frame, our model accurately detects the gaze target where the VAT method failed to do so. This demonstrates the superior performance of our model in terms of gaze target detection. In the second frame, our model accurately detected the speaker as the gaze target in a conversational scenario, while another model failed. Incorporating audio cues is crucial for gaze following, and audio-visual fusion can significantly improve accuracy, especially in real-world scenarios.}
\vspace{-0.3cm}
\label{fig:fig5}
\end{figure*}

\subsection{Data Collection}

The VideoGazeSpeech Dataset contains $29$ videos with audio information in mp4 format, and the main task targeted is gaze estimation in social situations. This dataset was selected from the video dataset with audio \cite{xu2018find}, and the original dataset only targets gaze estimation, which is an entirely different annotation and task from gaze following. Therefore, in this project, we need to re-annotate this dataset in its entirety. We used manual labelling in the labelling process, splitting the video by frame and labelling each gaze following each character object in each frame through the DarkLabel tool. To ensure the accuracy of the dataset, we also had three different reviewers check the dataset. 

We chose the tagged videos for each video to guarantee as much of an equitable distribution of data as possible in terms of the number of frames and persons in the movie. The average number of frames per video is evenly distributed in frames $400\sim500$, and the average number of persons in each video is evenly distributed in $2\sim4$.

\subsection{Data Processing}
In order to facilitate the training and adaptation of multiple types of neural networks and for the scalability of the database, this project also extends the VGS database into three formats: VOC format, COCO format, and VideoAttentionTarget format. 
The reader can directly utilize these datasets for method verification, eliminating the need for additional data transformation efforts. Moreover, the dataset is randomly partitioned into a training set and a test set in a 9:1 ratio, with the training set comprising $31,701$ frames and the test set encompassing $3,524$ frames.


\section{Experiments}
Our approach is novel in recognizing the significance of audio in gaze following, and as there are no existing methods directly comparable to our approach, we conducted a comprehensive evaluation of our gaze following model using our VGS dataset. We divided our experiments into two parts:  Comparison with SOTA methods and ablation with different backbones. In the ablation experiments, we examined the impact of different backbones (ResNet101, ResNet50, and ResNeXt-101) on the performance of our gaze candidate estimation model. 
In comparison experiments, we gauged our model against SOTA gaze detection models and investigated the impact of introducing multimodality on gaze target detection models in various ways.

Our experiments demonstrate that our proposed model outperforms other models. Specifically, the comparison experiments with different backbones inside the gaze candidate estimation model and with advanced gaze following algorithms highlight the efficacy of our multimodal processing approach that leverages audio-vision features. Our findings suggest that integrating audio and visual information can improve the performance of gaze following tasks.

\subsection{Evaluation Metrics}
AP focuses on the model's ability to cover positive samples and identify them. Suppose there are M positive cases in these N samples, then we get M recall values (1/M, 2/M, ... , M/M), and for each recall value r, we can calculate the maximum precision corresponding to ($r'>= r$), M/M), for each recall value r, the maximum precision corresponding to ($r' >= r$) can be calculated, and then average these M precision values to get the final AP value.  Considering this project only focuses on detecting gaze targets, there is only one target class, the gaze targets, and there is no need to calculate the mAP.

\subsection{Implementation Details}
The current gaze following models use raw frame and head position as their feature map \cite{chong2020detecting,recasens2015they,fang2021dual,hu2022we}. However, our proposed gaze candidate estimation model integrates audio information with video information for multimodal fusion training, which is a novel approach in the gazing field. In this experiment, We aim to explore the impact of multimodality on gaze tracking, comparing two variables: \textit{w/o} audio (visual cues only) and \textit{with} audio (audio and visual cues). Our goal is to examine the effect of audio-vision fusion on gaze following detection.

To conduct this experiment, we used the VGS database proposed in our project and adjusted the feature-processed data into coco format to train the gaze candidate estimation network. The gaze candidate estimation model employed ResNet-101, ResNet-50, ResNeXt-101, and FPN [54] as the neck of the gaze candidate estimation network, which reprocesses and rationalizes important features extracted from the backbone. During the training process, the learning rate was set to $0.0025$, the number of epochs was set to $12$. We used two NVIDIA RTX 3090 in this experiment.

\subsection{Quantitative Analysis}

\subsubsection{Comparison with SOTA methods}

In our experiment, we compared traditional CNN methods and Transformer methods in the context of gaze following. We used DETR \cite{zhu2021deformable} as the representative of the Transformer method due to its SOTA performance in computer vision. We also included VAT \cite{chong2020detecting}, a gaze following domain-based CNN model, to demonstrate the innovation and feasibility of our proposed gaze candidate estimation model.

Our gaze candidate estimation network is a multimodal network structure, so we explored the performance of different modalities in different network models to verify that the richer information brought by multimodality would be helpful for gaze following detection. During training, we used DETR to train our VGS database, and the VGS dataset was used in COCO format.

\begin{figure}[t]
\centering
\includegraphics[width=0.95\linewidth]{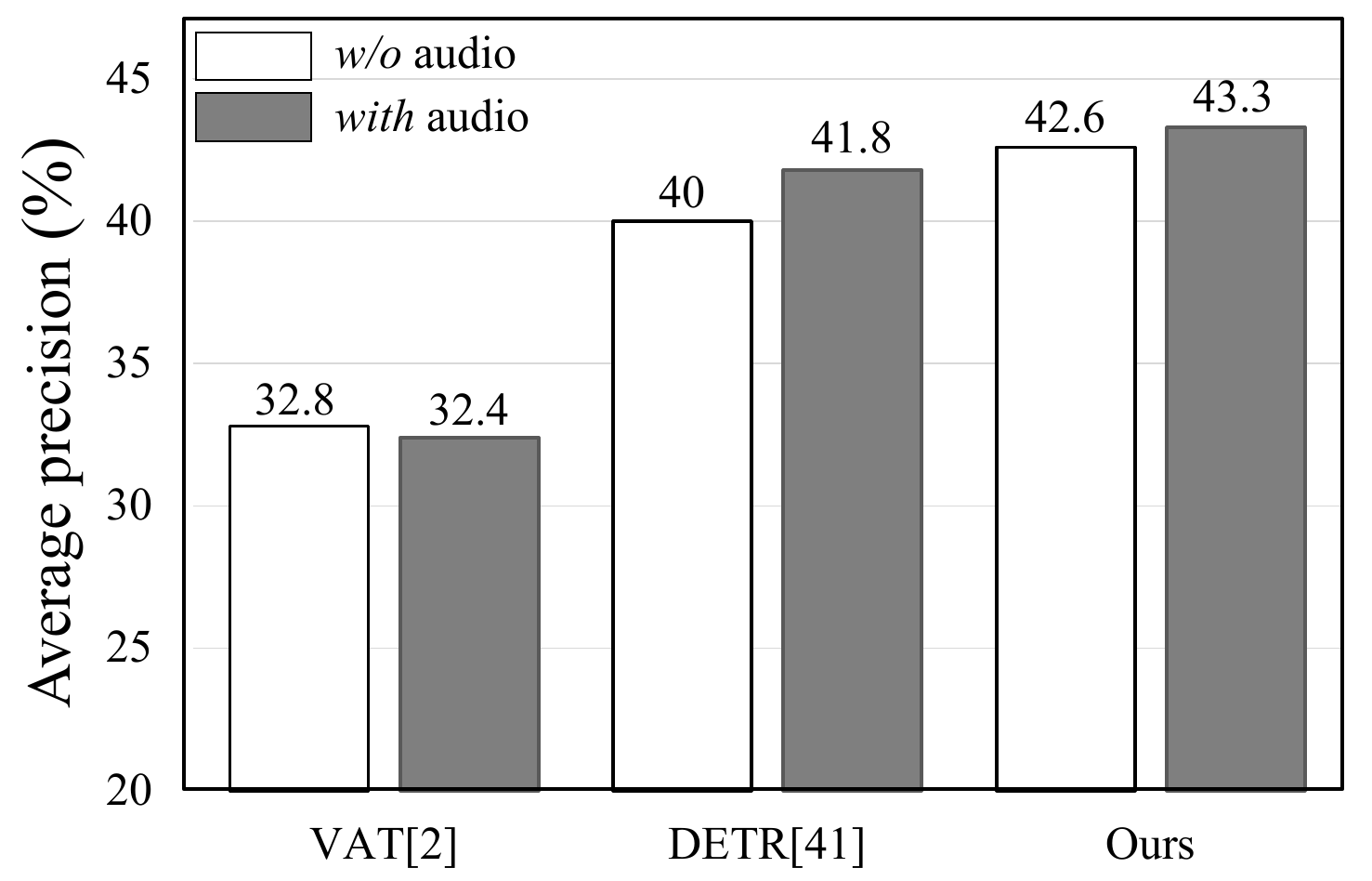}
\caption{Quantitative evaluation in comparison with state-of-the-art methods on our VGS dataset in the AP (Average precision) metric $\uparrow$ (higher is better). Our method outperforms DETR and VAT.}
\vspace{-0.3cm}
\label{fig:comparison}
\end{figure}

The results, shown in Fig. \ref{fig:comparison}, demonstrate that our multimodal network structure gaze candidate estimation network ($0.433$) outperforms DETR ($0.418$) and VAT ($0.324$) in terms of AP performance. Moreover, as the modality increases, the AP of our method and Transformer method performs better than that of a single modality. Interestingly, we found that VAT performs worse when audio cues are added to the feature map, indicating that its network is too simple to handle multimodal information.

These results suggest that incorporating audio information into gaze following models, as we did in our gaze candidate estimation model, can lead to significant improvements in accuracy, particularly in real-world scenarios where audio cues play a crucial role. The superiority of our multimodal network structure over traditional CNN methods and Transformer methods also highlights the importance of fusing multimodal information for gaze following detection.

\subsubsection{Different Backbones}

In this gaze candidate estimation work, we conducted experiments to compare and test three backbones: ResNet-101, ResNet-50 \cite{he2016deep}, and ResNeXt-101 \cite{xie2017aggregated}. Our results indicate that ResNeXt outperforms ResNet with the same number of parameters, which is consistent with Fig. \ref{fig:ablation}. Specifically, ResNeXt has a higher average precision (AP) than ResNet for different modal treatments. We also found that increasing the number of neural network layers from $50$ to $101$ for ResNet leads to a slight performance improvement, but not as significant as ResNeXt.

Notably, Fig. \ref{fig:ablation} shows that each backbone with audio-vision feature map outperforms the visual feature map, indicating that audio-vision fusion of feature can significantly improve the performance of gaze following target detection.

\subsection{Qualitative Analysis }
\begin{figure}[t]
\centering
\includegraphics[width=0.95\linewidth]{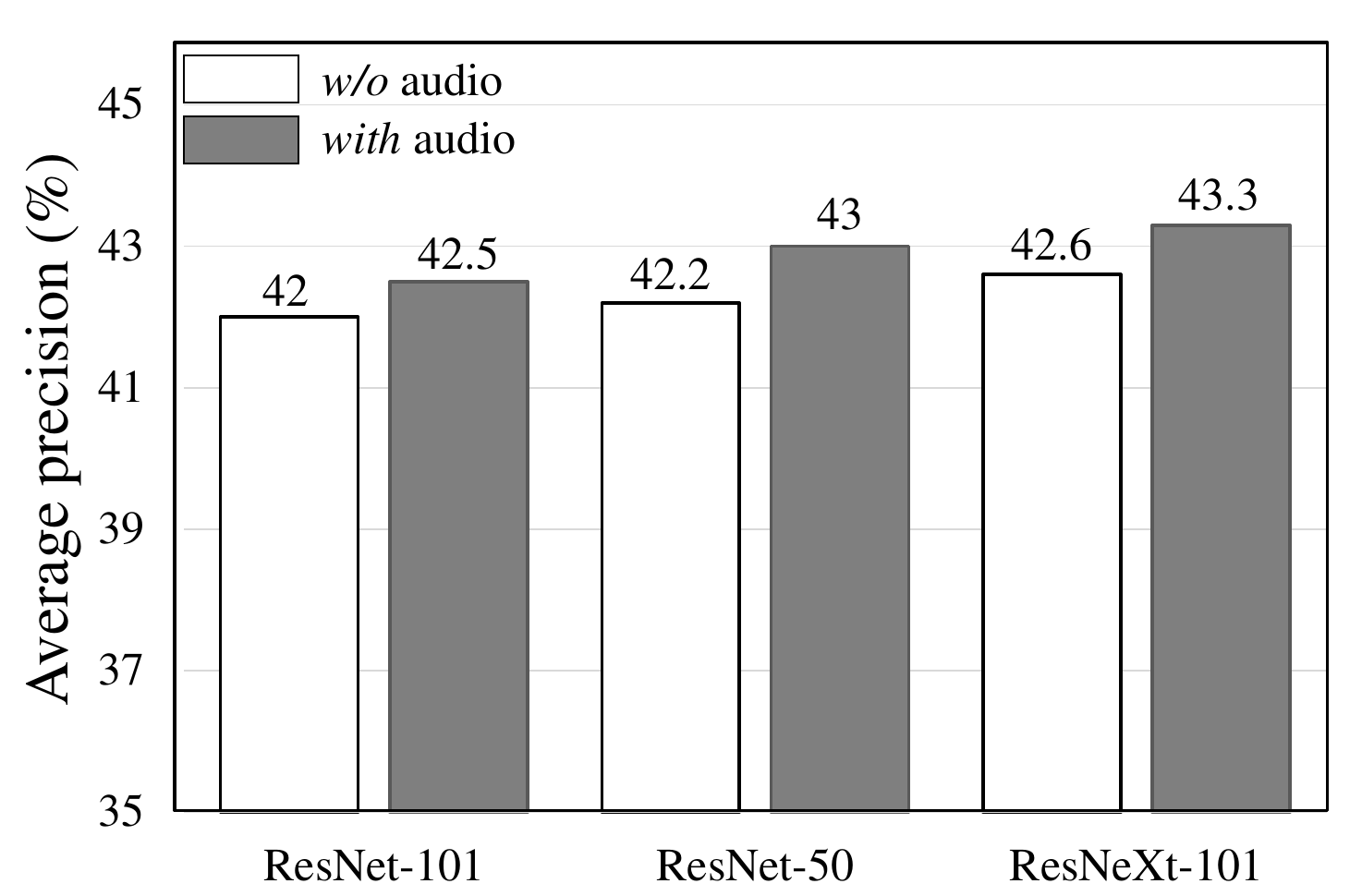}
\caption{Performance of Gaze Candidate Estimation Network with different backbones  in the AP (Average Precision) metric$\uparrow$ (higher is better)}
\vspace{-0.3cm}
\label{fig:ablation}
\end{figure}

We employed a rigorous evaluation approach to compare the performance of various gaze-following models using Gaussian heat maps generated from random samples of video data. The results in Fig. \ref{fig:ablation} and Fig. \ref{fig:comparison} clearly demonstrate the effectiveness of our multimodal model in enhancing the prediction accuracy of gaze-following models. In particular, the gaze candidate estimation model shows superior performance compared to other models. This is because the gaze candidate estimation model takes into account the speaker's mode, which improves its accuracy in social situations.

Furthermore, Fig.\ref{fig:fig5} illustrates how the gaze candidate estimation model's consideration of speaker mode can lead to more accurate analysis results. For example, in the second image of the first row, the person is looking at the speaker, whereas the VAT method wrongly detects the person looking at another listener. This finding underscores the importance of considering multimodal information in gaze-following models to achieve more robust and accurate results.



Our training results show the gaze candidate estimation network converges faster and more efficiently than the DETR model, which took sixfold time and eightfold epochs to converge, emphasizing our model's efficacy for social gaze-following. The superior performance of our multimodal model underlines the value of multimodal inputs in conversational gaze analysis, with considerable implications for advancing robust gaze behavior models in social contexts.

\section{Conclusion}

In this research project, we introduce a novel multimodal framework that overcomes the limitations of existing methods for gaze following in conversational settings. Our proposed approach leverages audio-vision fusion, which provides multiple sources of input and significantly improves detection accuracy and robustness. The framework learns subjects' identity information based on the correspondence of visual and audio features, while our gaze candidate estimation network leverages both identity information and scene images to estimate gaze candidates.

A major contribution of our study is the VideoGazeSpeech dataset, which includes annotated audio and video cues and is the first multimodal gaze tracking dataset. This dataset provides a valuable benchmark for evaluating the performance of gaze tracking models that utilize audio and video inputs. To evaluate the effectiveness of our approach, we conduct experiments on the VideoGazeSpeech dataset, demonstrating the advantage of audio-vision fusion. 

In conclusion, our proposed multimodal framework for gaze following in conversational settings and the VideoGazeSpeech dataset represent significant contributions to the field. It has the potential to enhance the accuracy and effectiveness of gaze following, ultimately improving human-robot interaction in conversational settings.

\bigskip 
\noindent\textbf{Acknowledgments:}This work was supported by the Institute of Information \& communications Technology Planning \& Evaluation (IITP) grant funded by the Korea government(MSIT) (No.2022-0-00608, Artificial intelligence research about multi-modal interactions for empathetic conversations with humans (50\%); No.2021-0-00034, Clustering technologies of fragmented data for time-based data analysis (50\%)). Yuqi Hou and Zhongqun Zhang were supported by China Scholarship Council Grant No.202308060328 and No.202208060266, respectively.

\newpage

{\small
\bibliographystyle{ieee_fullname}
\bibliography{egbib}
}

\end{document}

%% file: fig/teaser/item.tex
\begin{figure}
\centering
\includegraphics[width=1.0\linewidth]{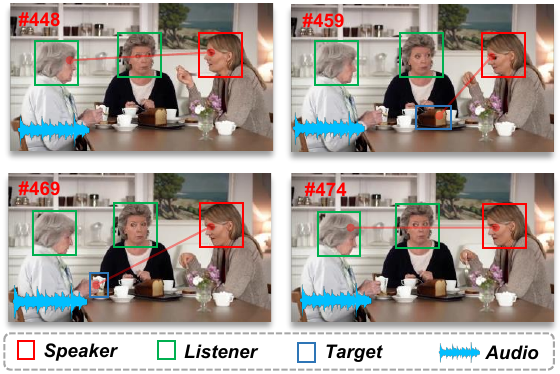}
\caption{We present a novel multimodal framework for conversational gaze following that utilizes audio-vision video input to generate accurate target detection for gaze following. Our approach produces annotated bounding boxes for the speaker, listener, and gaze target. To facilitate our methodology, we introduce VideoGazeSpeech (VGS) including annotated audio and video cues.}
\label{fig:teaser}
\end{figure}

%% file: tbl/quantitative/item2.tex
\newcolumntype{C}{>{\centering\arraybackslash}X}
\definecolor{mygreen}{RGB}{0,128,0}  

\begin{table}[t]

\caption{ Comparison of existing gaze following datasets. Our dataset is the first to provide audio modality which would encourage future research in multi-modal gaze following.}
\label{table1}
\small
\setlength{\tabcolsep}{5pt}
\renewcommand\arraystretch{1.2}
\resizebox{0.48\textwidth}{!}{
    \begin{tabular}{|l|c|c|c|c|}
     \hline
     \multirow{2}{*}{\textbf{Dataset}} &\multirow{2}{*}{\textbf{Pub.}}& \multirow{2}{*}{\textbf{Year}}  & \multicolumn{2}{c|}{\textbf{Modalities}} \\
     \cline{4-5}
     &&&\textbf{Vision}&\textbf{Audio} \\
      \hline
    GazeFollow \cite{recasens2015they}&NeurIPS&2015&\color{mygreen}{\cmark}&\textcolor{red}{\xmark}\\
    VideoGaze\cite{recasens2016following}&ICCV&2017&\color{mygreen}{\cmark}&\textcolor{red}{\textcolor{red}{\xmark}}\\
    VideoCoAtt \cite{fan2018inferring}&CVPR&2018&\color{mygreen}{\cmark}&\textcolor{red}{\textcolor{red}{\xmark}}\\
    Gaze360 \cite{kellnhofer2019gaze360}&ICCV&2019&\color{mygreen}{\cmark}&\textcolor{red}{\textcolor{red}{\xmark}}\\
    VideoAttentionTarget \cite{chong2020detecting}&CVPR&2020&\color{mygreen}{\cmark}&\textcolor{red}{\textcolor{red}{\xmark}}\\
    GazeFollow360 \cite{ Li_2021_ICCV}&ICCV&2021&\color{mygreen}{\cmark}&\textcolor{red}{\textcolor{red}{\xmark}}\\

     \hline
    Ours&  &2023&\color{mygreen}{\cmark}&\color{mygreen}{\cmark}\\

     \hline
    \end{tabular}
}
    
\vspace{-0.1cm}
\end{table}

%% file: fig/framework/item.tex
\begin{figure*}[ht]
\begin{center}
\centering
\includegraphics[width=\linewidth]{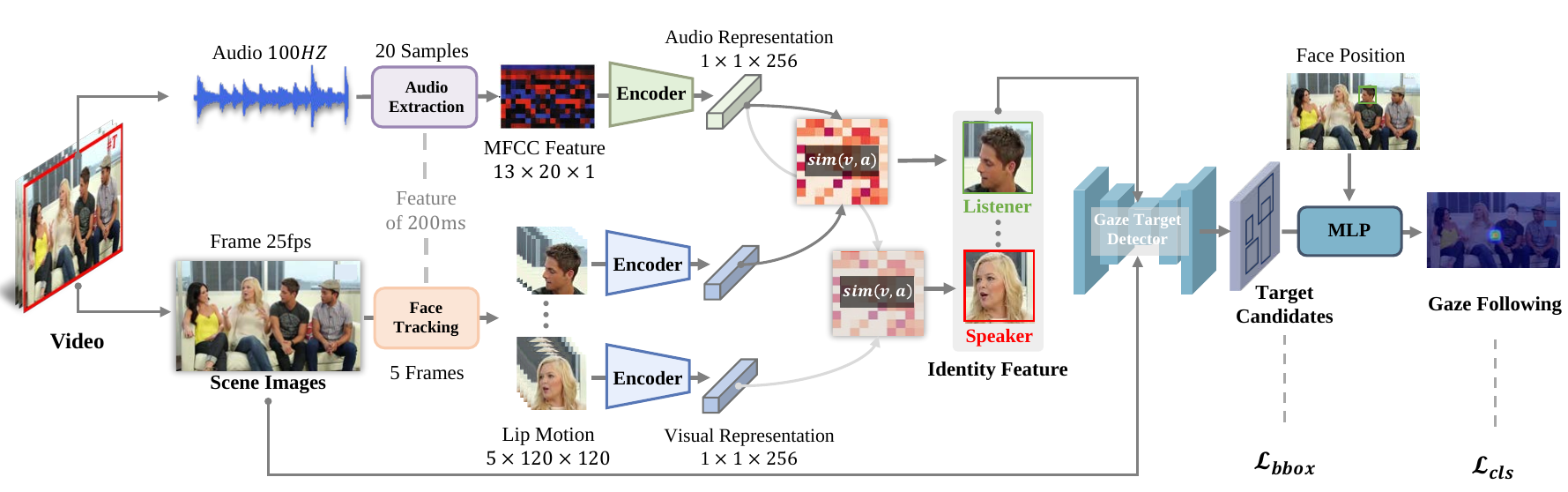}
\vspace{-0.5 cm}
\caption{MMGaze performs gaze following for each frame of videos. Given one frame and audio track, it first performs active speaker detection. MMGaze acquires audio feature of 200 ms (20 samples due to 100\textit{HZ}) near the timestamp of the given frame. It also acquires corresponding visual images of 200 ms (5 frames due to 25fps). Then, MMGaze extracts audio representation and visual representation corresponding to lip motion of each individual via SyncNet~\cite{chung2016out}. It computes the similarity between audio representation and visual representation of each individual, and distinguishes identity information.
MMGaze provides a gaze candidate estimation network. It contains a gaze target detector to estimate gaze target candidates from scene images enhanced by identity information. One multilayer perceptron (MLP) is used to predict the relationships between each subject and all candidates.  We select the candidate with the highest probability as the final gaze target for each subject.  
}

\label{framework}
\end{center}
\end{figure*}

%% file: PaperForReview.bbl
\begin{thebibliography}{10}\itemsep=-1pt

\bibitem{Bao_2020_ICPR}
Yiwei Bao, Yihua Cheng, Yunfei Liu, and Feng Lu.
\newblock Adaptive feature fusion network for gaze tracking in mobile tablets.
\newblock In {\em International Conference on Pattern Recognition (ICPR)}, 2020.

\bibitem{boccignone2020gaze}
Giuseppe Boccignone, Vittorio Cuculo, Alessandro D’Amelio, Giuliano Grossi, and Raffaella Lanzarotti.
\newblock On gaze deployment to audio-visual cues of social interactions.
\newblock {\em IEEE Access}, 8:161630--161654, 2020.

\bibitem{cheng_2022_aaai}
Yihua Cheng, Yiwei Bao, and Feng Lu.
\newblock Puregaze: Purifying gaze feature for generalizable gaze estimation.
\newblock In {\em AAAI}, 2022.

\bibitem{Cheng_2020_AAAI}
Yihua Cheng, Shiyao Huang, Fei Wang, Chen Qian, and Feng Lu.
\newblock A coarse-to-fine adaptive network for appearance-based gaze estimation.
\newblock In {\em Proceedings of the AAAI Conference on Artificial Intelligence}, 2020.

\bibitem{cheng2022icpr}
Yihua Cheng and Feng Lu.
\newblock Gaze estimation using transformer.
\newblock In {\em International Conference on Pattern Recognition (ICPR)}, 2022.

\bibitem{Cheng_2023_ICCV}
Yihua Cheng and Feng Lu.
\newblock Dvgaze: Dual-view gaze estimation.
\newblock In {\em Proceedings of the IEEE/CVF International Conference on Computer Vision (ICCV)}, pages 20632--20641, October 2023.

\bibitem{Cheng_2018_ECCV}
Yihua Cheng, Feng Lu, and Xucong Zhang.
\newblock Appearance-based gaze estimation via evaluation-guided asymmetric regression.
\newblock In {\em The European Conference on Computer Vision}, 2018.

\bibitem{Cheng_2021_arxiv}
Yihua Cheng, Haofei Wang, Yiwei Bao, and Feng Lu.
\newblock Appearance-based gaze estimation with deep learning: A review and benchmark.
\newblock {\em arXiv preprint arXiv:2104.12668}, 2021.

\bibitem{chong2020detecting}
Eunji Chong, Yongxin Wang, Nataniel Ruiz, and James~M Rehg.
\newblock Detecting attended visual targets in video.
\newblock In {\em Proceedings of the IEEE/CVF conference on computer vision and pattern recognition}, pages 5396--5406, 2020.

\bibitem{chung2016out}
Joon~Son Chung and Andrew Zisserman.
\newblock Out of time: automated lip sync in the wild.
\newblock In {\em Asian conference on computer vision}, pages 251--263. Springer, 2016.

\bibitem{d2021gazing}
Alessandro D'Amelio and Giuseppe Boccignone.
\newblock Gazing at social interactions between foraging and decision theory.
\newblock {\em Frontiers in neurorobotics}, 15:31, 2021.

\bibitem{davis1980comparison}
Steven Davis and Paul Mermelstein.
\newblock Comparison of parametric representations for monosyllabic word recognition in continuously spoken sentences.
\newblock {\em IEEE transactions on acoustics, speech, and signal processing}, 28(4):357--366, 1980.

\bibitem{dosovitskiy2020image}
Alexey Dosovitskiy, Lucas Beyer, Alexander Kolesnikov, Dirk Weissenborn, Xiaohua Zhai, Thomas Unterthiner, Mostafa Dehghani, Matthias Minderer, Georg Heigold, Sylvain Gelly, Jakob Uszkoreit, and Neil Houlsby.
\newblock An image is worth 16x16 words: Transformers for image recognition at scale.
\newblock In {\em International Conference on Learning Representations}, 2021.

\bibitem{doyle2001identification}
Melanie~C Doyle and Robert~J Snowden.
\newblock Identification of visual stimuli is improved by accompanying auditory stimuli: The role of eye movements and sound location.
\newblock {\em Perception}, 30(7):795--810, 2001.

\bibitem{epstein2020oops}
Dave Epstein, Boyuan Chen, and Carl Vondrick.
\newblock Oops! predicting unintentional action in video.
\newblock In {\em Proceedings of the IEEE/CVF Conference on Computer Vision and Pattern Recognition (CVPR)}, June 2020.

\bibitem{fan2018inferring}
Lifeng Fan, Yixin Chen, Ping Wei, Wenguan Wang, and Song-Chun Zhu.
\newblock Inferring shared attention in social scene videos.
\newblock In {\em Proceedings of the IEEE Conference on Computer Vision and Pattern Recognition}, pages 6460--6468, 2018.

\bibitem{fan2019understanding}
Lifeng Fan, Wenguan Wang, Siyuan Huang, Xinyu Tang, and Song-Chun Zhu.
\newblock Understanding human gaze communication by spatio-temporal graph reasoning.
\newblock In {\em Proceedings of the IEEE International Conference on Computer Vision}, pages 5724--5733, 2019.

\bibitem{fang2021dual}
Yi Fang, Jiapeng Tang, Wang Shen, Wei Shen, Xiao Gu, Li Song, and Guangtao Zhai.
\newblock Dual attention guided gaze target detection in the wild.
\newblock In {\em Proceedings of the IEEE/CVF conference on computer vision and pattern recognition}, pages 11390--11399, 2021.

\bibitem{gao2020survey}
Jing Gao, Peng Li, Zhikui Chen, and Jianing Zhang.
\newblock A survey on deep learning for multimodal data fusion.
\newblock {\em Neural Computation}, 32(5):829--864, 2020.

\bibitem{he2017mask}
Kaiming He, Georgia Gkioxari, Piotr Doll{\'a}r, and Ross Girshick.
\newblock Mask r-cnn.
\newblock In {\em Proceedings of the IEEE/CVF International Conference on Computer Vision (ICCV)}, pages 2961--2969, 2017.

\bibitem{he2016deep}
Kaiming He, Xiangyu Zhang, Shaoqing Ren, and Jian Sun.
\newblock Deep residual learning for image recognition.
\newblock In {\em Proceedings of the IEEE conference on computer vision and pattern recognition}, pages 770--778, 2016.

\bibitem{horanyi2023g}
Nora Horanyi, Yuqi Hou, Ales Leonardis, and Hyung~Jin Chang.
\newblock G-daic: A gaze initialized framework for description and aesthetic-based image cropping.
\newblock {\em Proceedings of the ACM on Human-Computer Interaction}, 7(ETRA):1--19, 2023.

\bibitem{Horanyi_2023_CVPR}
Nora Horanyi, Linfang Zheng, Eunji Chong, Ale\v{s} Leonardis, and Hyung~Jin Chang.
\newblock Where are they looking in the 3d space?
\newblock In {\em Proceedings of the IEEE/CVF Conference on Computer Vision and Pattern Recognition (CVPR) Workshops}, pages 2678--2687, June 2023.

\bibitem{hu2022we}
Zhengxi Hu, Dingye Yang, Shilei Cheng, Lei Zhou, Shichao Wu, and Jingtai Liu.
\newblock We know where they are looking at from the rgb-d camera: Gaze following in 3d.
\newblock {\em IEEE Transactions on Instrumentation and Measurement}, 71:1--14, 2022.

\bibitem{jia2021scaling}
Chao Jia, Yinfei Yang, Ye Xia, Yi-Ting Chen, Zarana Parekh, Hieu Pham, Quoc Le, Yun-Hsuan Sung, Zhen Li, and Tom Duerig.
\newblock Scaling up visual and vision-language representation learning with noisy text supervision.
\newblock In {\em International Conference on Machine Learning}, pages 4904--4916. PMLR, 2021.

\bibitem{kellnhofer2019gaze360}
Petr Kellnhofer, Adria Recasens, Simon Stent, Wojciech Matusik, and Antonio Torralba.
\newblock Gaze360: Physically unconstrained gaze estimation in the wild.
\newblock In {\em Proceedings of the IEEE/CVF International Conference on Computer Vision}, pages 6912--6921, 2019.

\bibitem{Li_2021_ICCV}
Yunhao Li, Wei Shen, Zhongpai Gao, Yucheng Zhu, Guangtao Zhai, and Guodong Guo.
\newblock Looking here or there? gaze following in 360-degree images.
\newblock In {\em Proceedings of the IEEE/CVF International Conference on Computer Vision (ICCV)}, pages 3742--3751, October 2021.

\bibitem{marin2014detecting}
Manuel~Jes{\'u}s Marin-Jimenez, Andrew Zisserman, Marcin Eichner, and Vittorio Ferrari.
\newblock Detecting people looking at each other in videos.
\newblock {\em International Journal of Computer Vision}, 106(3):282--296, 2014.

\bibitem{masse2017tracking}
Beno{\^\i}t Mass{\'e}, Sil{\`e}ye Ba, and Radu Horaud.
\newblock Tracking gaze and visual focus of attention of people involved in social interaction.
\newblock {\em IEEE transactions on pattern analysis and machine intelligence}, 40(11):2711--2724, 2017.

\bibitem{mcmillan2019designing}
Donald McMillan, Barry Brown, Ikkaku Kawaguchi, Razan Jaber, Jordi Solsona~Belenguer, and Hideaki Kuzuoka.
\newblock Designing with gaze: Tama--a gaze activated smart-speaker.
\newblock {\em Proceedings of the ACM on Human-Computer Interaction}, 3(CSCW):1--26, 2019.

\bibitem{min2014sound}
Xiongkuo Min, Guangtao Zhai, Zhongpai Gao, Chunjia Hu, and Xiaokang Yang.
\newblock Sound influences visual attention discriminately in videos.
\newblock In {\em 2014 Sixth International Workshop on Quality of Multimedia Experience (QoMEX)}, pages 153--158. IEEE, 2014.

\bibitem{Nonaka_2022_CVPR}
Soma Nonaka, Shohei Nobuhara, and Ko Nishino.
\newblock Dynamic 3d gaze from afar: Deep gaze estimation from temporal eye-head-body coordination.
\newblock In {\em Proceedings of the IEEE/CVF Conference on Computer Vision and Pattern Recognition (CVPR)}, pages 2192--2201, June 2022.

\bibitem{perrott1990auditory}
David~R Perrott, Kourosh Saberi, Kathleen Brown, and Thomas~Z Strybel.
\newblock Auditory psychomotor coordination and visual search performance.
\newblock {\em Perception \& psychophysics}, 48(3):214--226, 1990.

\bibitem{radford2021learning}
Alec Radford, Jong~Wook Kim, Chris Hallacy, Aditya Ramesh, Gabriel Goh, Sandhini Agarwal, Girish Sastry, Amanda Askell, Pamela Mishkin, Jack Clark, et~al.
\newblock Learning transferable visual models from natural language supervision.
\newblock In {\em International Conference on Machine Learning}, pages 8748--8763. PMLR, 2021.

\bibitem{recasens2015they}
Adria Recasens, Aditya Khosla, Carl Vondrick, and Antonio Torralba.
\newblock Where are they looking?
\newblock {\em Advances in neural information processing systems}, 28, 2015.

\bibitem{recasens2016following}
Adri{\`a} Recasens, Carl Vondrick, Aditya Khosla, and Antonio Torralba.
\newblock Following gaze across views.
\newblock {\em arXiv preprint arXiv:1612.03094}, 2016.

\bibitem{sumer2020attention}
Omer Sumer, Peter Gerjets, Ulrich Trautwein, and Enkelejda Kasneci.
\newblock Attention flow: End-to-end joint attention estimation.
\newblock In {\em The IEEE Winter Conference on Applications of Computer Vision}, pages 3327--3336, 2020.

\bibitem{tavakoli2019dave}
Hamed~Rezazadegan Tavakoli, Ali Borji, Juho Kannala, and Esa Rahtu.
\newblock Deep audio-visual saliency: Baseline model and data.
\newblock In {\em ACM Symposium on Eye Tracking Research and Applications}, ETRA '20 Short Papers, New York, NY, USA, 2020. Association for Computing Machinery.

\bibitem{tu2022end}
Danyang Tu, Xiongkuo Min, Huiyu Duan, Guodong Guo, Guangtao Zhai, and Wei Shen.
\newblock End-to-end human-gaze-target detection with transformers.
\newblock In {\em 2022 IEEE/CVF Conference on Computer Vision and Pattern Recognition (CVPR)}, pages 2192--2200. IEEE, 2022.

\bibitem{van2008audiovisual}
Erik Van~der Burg, Christian~NL Olivers, Adelbert~W Bronkhorst, and Jan Theeuwes.
\newblock Audiovisual events capture attention: Evidence from temporal order judgments.
\newblock {\em Journal of vision}, 8(5):2--2, 2008.

\bibitem{van2008pip}
Erik Van~der Burg, Christian~NL Olivers, Adelbert~W Bronkhorst, and Jan Theeuwes.
\newblock Pip and pop: nonspatial auditory signals improve spatial visual search.
\newblock {\em Journal of Experimental Psychology: Human Perception and Performance}, 34(5):1053, 2008.

\bibitem{vasudevan2018object}
Arun~Balajee Vasudevan, Dengxin Dai, and Luc Van~Gool.
\newblock Object referring in videos with language and human gaze.
\newblock In {\em Proceedings of the IEEE Conference on Computer Vision and Pattern Recognition}, pages 4129--4138, 2018.

\bibitem{vroomen2000sound}
Jean Vroomen and Beatrice~de Gelder.
\newblock Sound enhances visual perception: cross-modal effects of auditory organization on vision.
\newblock {\em Journal of experimental psychology: Human perception and performance}, 26(5):1583, 2000.

\bibitem{wei2018looking}
Ping Wei, Yang Liu, Tianmin Shu, Nanning Zheng, and Song-Chun Zhu.
\newblock Where and why are they looking? jointly inferring human attention and intentions in complex tasks.
\newblock In {\em Proceedings of the IEEE Conference on Computer Vision and Pattern Recognition}, pages 6801--6809, 2018.

\bibitem{xie2017aggregated}
Saining Xie, Ross Girshick, Piotr Doll{\'a}r, Zhuowen Tu, and Kaiming He.
\newblock Aggregated residual transformations for deep neural networks.
\newblock In {\em Proceedings of the IEEE conference on computer vision and pattern recognition}, pages 1492--1500, 2017.

\bibitem{xu2018find}
Mai Xu, Yufan Liu, Roland Hu, and Feng He.
\newblock Find who to look at: Turning from action to saliency.
\newblock {\em IEEE Transactions on Image Processing}, 27(9):4529--4544, 2018.

\bibitem{zhang2017s3fd}
Shifeng Zhang, Xiangyu Zhu, Zhen Lei, Hailin Shi, Xiaobo Wang, and Stan~Z Li.
\newblock S3fd: Single shot scale-invariant face detector.
\newblock In {\em Proceedings of the IEEE international conference on computer vision}, pages 192--201, 2017.

\bibitem{zhu2021deformable}
Xizhou Zhu, Weijie Su, Lewei Lu, Bin Li, Xiaogang Wang, and Jifeng Dai.
\newblock Deformable {\{}detr{\}}: Deformable transformers for end-to-end object detection.
\newblock In {\em International Conference on Learning Representations}, 2021.

\end{thebibliography}
